\documentclass[acmtog]{acmart}
\acmSubmissionID{}

\usepackage{booktabs}
\usepackage[ruled]{algorithm2e} 

\SetAlFnt{\small}
\SetAlCapFnt{\small}
\SetAlCapNameFnt{\small}
\SetAlCapHSkip{0pt}

\usepackage{tabularray}
\usepackage{graphicx}
\usepackage{subcaption}

\SetAlFnt{\small}
\SetAlCapFnt{\small}
\SetAlCapNameFnt{\small}
\SetAlCapHSkip{0pt}

\renewcommand\footnotetextcopyrightpermission[1]{} 

\usepackage{xcolor}  
\usepackage{xspace}  
\usepackage{cleveref}

\usepackage{multirow}
\usepackage{tabularx} 
\usepackage{graphicx} 
\usepackage{float}
\usepackage{colortbl} 
\usepackage{booktabs} 

\usepackage{amsmath}

\definecolor{cvprblue}{rgb}{0.21,0.49,0.74}
\definecolor{ForestGreen}{rgb}{0, 0.69, 0.31}
\definecolor{00purp}{RGB}{114, 58, 107}

\newcommand{\methodname}{\textbf{\textcolor{00purp}{RelightVid}}\xspace}

\begin{document}

\title{\textcolor{00purp}{RelightVid}: Temporal-Consistent Diffusion Model for Video Relighting}

\author{Ye Fang$^{1,2\,*}$ \quad Zeyi Sun$^{1,3\,*}$ \quad Shangzhan Zhang$^4$ \quad Tong Wu$^{5}$ \quad Yinghao Xu$^5$ \\
Pan Zhang$^1$ \quad Jiaqi Wang$^1$ \quad Gordon Wetzstein$^5$ \quad Dahua Lin$^{1,6}$}
\affiliation{
\institution{\\
$^1$Shanghai AI Laboratory \quad $^2$Fudan University \quad $^3$Shanghai Jiao Tong University \quad $^4$Zhejiang University\\ \quad $^5$Stanford University \quad $^6$The Chinese University of Hong Kong}
}

\begin{abstract}
\renewcommand{\thefootnote}{\fnsymbol{footnote}}  

\begin{flushright}
    \footnotetext[1]{Equal contribution.}
\end{flushright}

\vspace{-2mm}
Diffusion models have demonstrated remarkable success in image generation and editing, with recent advancements enabling albedo-preserving image relighting. However, applying these models to video relighting remains challenging due to the lack of paired video relighting datasets and the high demands for output fidelity and temporal consistency, further complicated by the inherent randomness of diffusion models. To address these challenges, we introduce \methodname, a flexible framework for video relighting that can accept background video, text prompts, or environment maps as relighting conditions. Trained on in-the-wild videos with carefully designed illumination augmentations and rendered videos under extreme dynamic lighting, \methodname achieves arbitrary video relighting with high temporal consistency without intrinsic decomposition while preserving the illumination priors of its image backbone.
\end{abstract}



\begin{teaserfigure}
    \centering
    \includegraphics[width=1.00\textwidth]{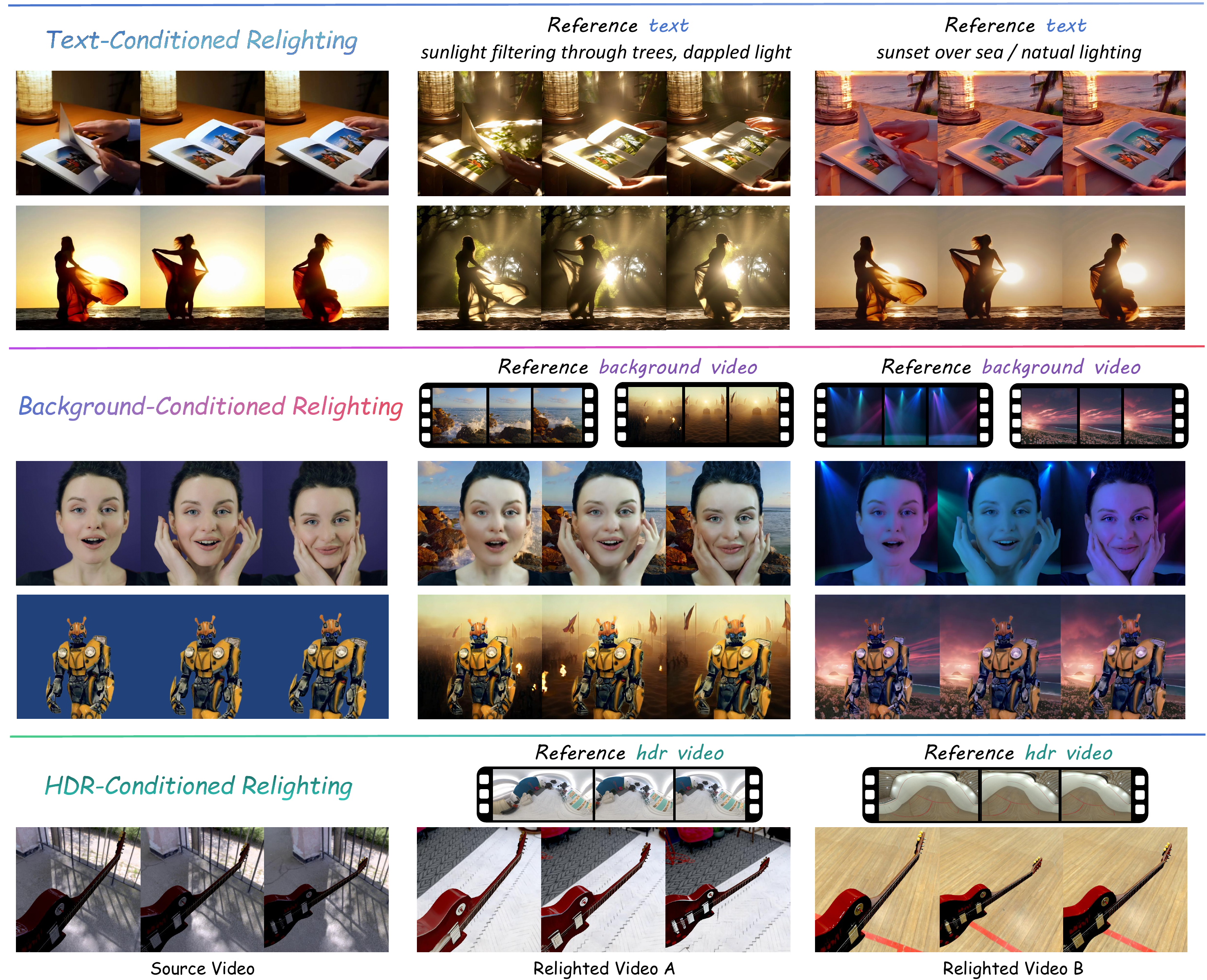}
    \vspace{-8mm}
    \caption{\methodname can perform high quality video relighting given a single video as input conditioned on text, background video and HDR environment video.}
    \label{fig:teaser}
\end{teaserfigure}

\maketitle

\vspace{-2mm}
\section{Introduction}
\label{sec:intro}


Lighting and its interactions with portraits and objects form the cornerstone of vision and imaging, shaping how we perceive both the physical world and its digital representations. The ability to relight a video—modifying the illumination of foreground subjects as if captured under different lighting scenarios—holds immense potential across various domains such as filmmaking~\cite{richardt2012coherent}, gaming, and augmented reality~\cite{debevec2008rendering, li2022physically}. By precisely controlling lighting, creators can not only enhance the visual experience but also gain greater artistic flexibility to meet the diverse demands of scenes.

Relighting dynamic foreground subjects in video under varying lighting conditions remains a significant challenge, particularly in maintaining temporal consistency and realistic lighting interactions. While inverse rendering~\cite{xia2016recovering,nam2018practical, zhang2021physg, cai2024real}  can decompose intrinsic properties and lighting, it relies on complex inputs like HDR images~\cite{reinhard2020high} or SH coefficients~\cite{ramamoorthi2001efficient} for accurate relighting. In practical scenarios, however, users often prefer simpler input, such as textual guidance or reference background videos as conditions, which limits the applicability of the above methods. Additionally, these techniques struggle with generalization, typically limited to portrait or simple object relighting, and fail to model lighting effectively in complex dynamic scenarios.

Recent advancements in diffusion models~\cite{dhariwal2021diffusion,ho2020denoising,song2020denoising,blattmann2023stable} trained on large-scale image and video datasets have demonstrated the ability to learn essential dynamics and physical priors. This enables them to perform physical rendering effects without explicitly relying on traditional physical modeling. Specifically, a growing body of works~\cite{ren2024relightful,jin2024neural,iclight} focuses on fine-tuning pre-trained diffusion models for tasks such as single-image relighting or illumination manipulation.
Notably, IC-Light~\cite{iclight} has emerged as a prominent approach, leveraging high-quality synthetic data and a consistent lighting loss function to achieve albedo-preserving relighting. However, extending such image-based techniques to videos introduces significant challenges. 
For example, a direct approach is applying IC-Light on a per-frame basis, but leads to substantial temporal inconsistencies. This stems from the inherent uncertainty in generative models, where the same input can yield multiple plausible outputs.
Furthermore, the scarcity of real or synthetic video relighting datasets presents another challenge. These datasets are crucial for fine-tuning models, as they enable the model to learn both the temporal consistency of illumination and the priors for complex dynamic light interactions, ensuring the naturalness and realism of generated relighting video.
To address these challenges, we introduce \methodname, a flexible framework lifting the capabilities of IC-Light~\cite{iclight} to video relighting. 
First, to overcome the issue of limited data, we introduce LightAtlas, a comprehensive video dataset created through a carefully designed augmentation pipeline. This dataset includes a large collection of real-world video footage and 3D-rendered data, along with corresponding lighting conditions and augmented pairs, providing the model with a rich prior knowledge of lighting in videos. 
Second, to tackle temporal consistency, we incorporate temporal layers into our model. These layers capture temporal dependencies between frames, ensuring high-quality relighting with strong temporal consistency, while maintaining the albedo-preserving capability of IC-Light. Finally, to enhance applicability and compatibility with varying lighting conditions, we support diverse types of inputs, including background videos, texture prompts, and precise HDR environment maps. 

Comprehensive experimental results demonstrate that our approach achieves high-quality, temporally consistent video relighting under multi-modal conditions, significantly outperforming the baseline in both qualitative and quantitative comparisons.
We believe \methodname can serve as a versatile tool for video relighting on arbitrary foreground subjects and pave the way for the application of video diffusion models in reverse rendering and generative tasks within the field of graphics. 

\begin{figure*}[t]
\centering
\includegraphics[width=0.95\textwidth]{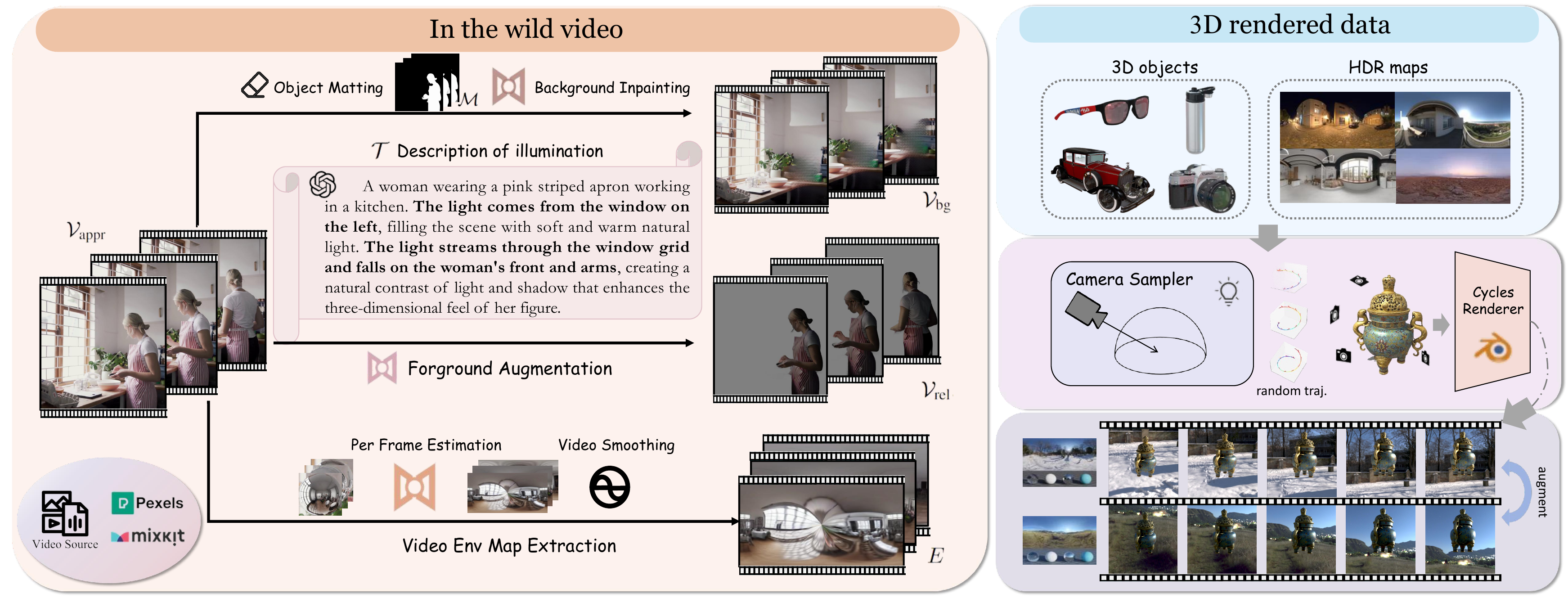}
\vspace{-4mm}
\caption{\textbf{LightAtlas Data Pipeline} generates high quality video relighting pairs both based on in the wild videos and 3D rendered data.}
\vspace{-4mm}
\label{fig:data}
\end{figure*}

\section{Related Work}

\noindent \textbf{Diffusion Models for Illumination Editing.} Recent  advancement of text-to-image diffusion models~\cite{rombach2022high, dhariwal2021diffusion, ho2020denoising, song2020denoising} have demonstrated strong capabilities in learning real-world image priors. Fine-tuned versions of these models have been successfully applied to a wide array of tasks, including image editing~\cite{alaluf2024cross,couairon2022diffedit,hertz2022prompt,brooks2023instructpix2pix,sun2024x}, geometric prediction~\cite{ke2024repurposing,fu2025geowizard}, 3D generation~\cite{anciukevivcius2023renderdiffusion,sun2024bootstrap3d,chen2023single,tang2023make,poole2022dreamfusion} and more recently, illumination manipulation~\cite{ren2024relightful,deng2025flashtex,zeng2024dilightnet,kocsis2024lightit,iclight}. Directly applying these models to videos via per-frame relighting often results in significant temporal inconsistencies due to the inherent ambiguity of lighting conditions. In this work, we propose extending IC-Light~\cite{iclight} into a video relighting model, which supports more flexible control based on background video, text, or full environment maps. Through a meticulously designed training pipeline, we achieve high-quality video relighting with enhanced temporal consistency, while preserving the lighting priors learned by IC-Light.

\noindent \textbf{Video Editing and Video Diffusion Models.} Recent years have seen remarkable breakthrough in video diffusion models~\cite{blattmann2023stable,guo2023animatediff,videoworldsimulators2024,yang2024cogvideox,xing2025dynamicrafter}. These models, after learning on real world videos, can generate high quality videos that obey real world physical laws, including illuminations. There are also works done by leveraging these models to achieve general video editing in both training~\cite{cheng2023consistent,singer2025video,mou2024revideo} and training free~\cite{ku2024anyv2v,ling2024motionclone,bu2024broadway} ways. While these methods excel in general video editing, achieving high-quality video relighting that preserves critical details such as albedo, lighting consistency, and scene realism remains a significant challenge. Our work represents an early exploration into leveraging video diffusion models to enable high-quality video relighting.

\noindent \textbf{Video Relighting.}
Current video relighting methods primarily focus on portrait videos. \cite{zhang2021neural} propose a neural approach for consistent relighting using a hybrid encoder-decoder with lighting disentanglement and temporal modeling.
\cite{kim2024switchlight} achieves relighting by explicitly predicting normal and shading maps and requiring HDR environment maps as input.
\cite{cai2024real} use NeRF to model the portrait head, achieving high-quality and real-time relighting.
These approaches are limited to portraits and rely heavily on explicit inputs, while our work enables relighting for arbitrary subjects with diverse and more practical conditions from a single input video.
\section{Methods}

\begin{figure*}[t]
\centering
\includegraphics[width=0.95\textwidth]{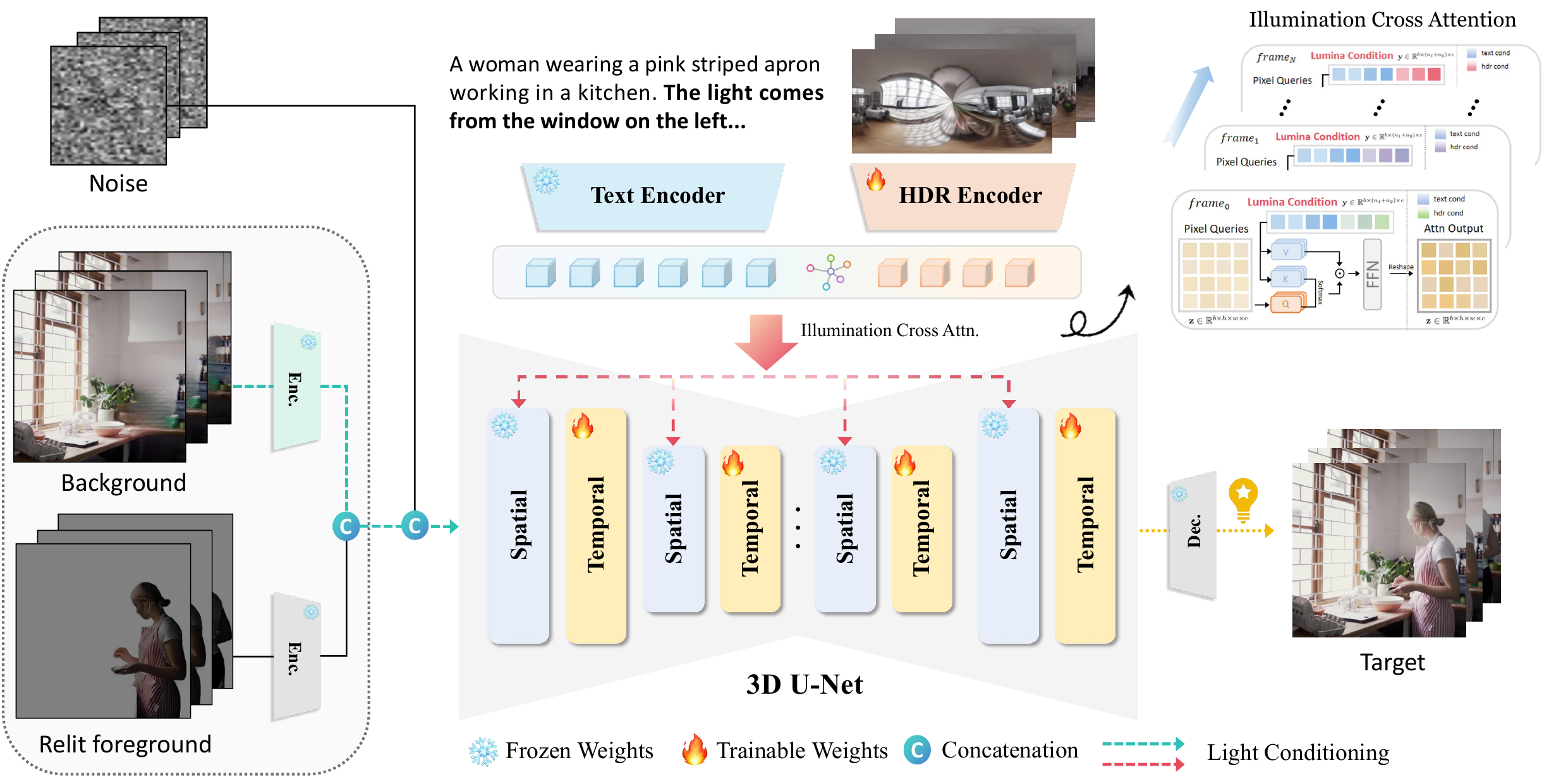}
\vspace{-3mm}
\caption{\textbf{Model Design} to lift image diffusion model for temporal consistent video relighting under text prompt, background video and HDR video map.}
\vspace{-4mm}
\label{fig:pipe}
\end{figure*}



We propose an efficient method named \methodname for editing the illumination in videos with consistent temporal performance. In \Cref{sec3.1}, we first introduce LightAtlas, a high quality video relighting dataset constructed from real video and 3D data renderings. We further present the design and training framework of \methodname in \Cref{sec3.2}, which supports temporally consistent video relighting under multi-modal conditions. 

\subsection{LightAtlas Data Collection Pipeline}
\label{sec3.1}
Training a model for arbitrary video-to-video illumination editing heavily depends on the availability of large-scale paired data. Due to the rarity of extreme dynamic lighting conditions in real-world videos, we utilize both in-the-wild video data to preserve photorealism and 3D-rendered data to effectively augment training under extreme lighting scenarios, as illustrated in \cref{fig:data}.
Each appearance video $\mathcal{V}_{\text{appr}} \in \mathbb{R}^{f \times h \times w \times 3}$ is paired with five types of augmented data to facilitate illumination modeling and learning: 
\[
\mathcal{V}_{\text{appr}} \leftrightarrow \{ \mathcal{V}_{\text{rel}}, \mathcal{V}_{\text{bg}}, E, \mathcal{T}, \mathcal{M} \},
\]
where $\mathcal{V}_{\text{rel}} \in \mathbb{R}^{f \times h \times w \times 3}$ represents the relit foreground video, $\mathcal{V}_{\text{bg}} \in \mathbb{R}^{f \times h \times w \times 3}$ is the background video, $E \in \mathbb{R}^{f \times 32 \times 32 \times 3}$ denotes the convoluted temporal environment map, $\mathcal{T}$ is the caption describing illumination changes, and $\mathcal{M} \in \mathbb{R}^{f \times h \times w}$ represents the foreground mask.



\subsubsection{In-the-wild video.} 
Generating paired data for in-the-wild videos poses significant challenges due to the complexity of obtaining high-quality and consistent illumination conditions. For real-world appearance videos $\mathcal{V}_{\text{appr}}$, we apply a 2D image relighting method (e.g., IC-Light~\cite{iclight}) frame-by-frame to generate augmented relit foreground videos $\mathcal{V}_{\text{rel}}$ under different illumination settings. To extract the object’s foreground mask, we leverage the powerful object matting tool InSPyReNet~\cite{kim2022revisiting}, while the inpainted background videos are obtained using video inpainting tool ProPainter~\cite{zhou2023propainter}. High Dynamic Range (HDR) environment maps for each frames are extracted from the video using DiffusionLight~\cite{phongthawee2024diffusionlight}, and then smoothed through temporal convolution. Additionally, we use GPT-4V~\cite{openai2023gpt} to generate captions describing the video (focusing on environmental and lighting details) and further filter the captions to retain 20K high-quality meta videos. 

Among augmented pairs, the background video, environment map, and caption are utilized as conditions for illumination modeling, with $\mathcal{V}_{\text{rel}}$ serving as the model input and $\mathcal{V}_{\text{appr}}$ as the target output. Since the target video is derived from real-world scenarios, it allows the model to learn the real data distribution and the temporal coherence across frames. Although this part of the data is of high photorealism, some input illumination conditions, particularly the environment map (HDR), include noise through estimations. To enhance the precise condition of HDR video, we incorporate auxiliary training data from 3D render engine to provide more precise control and improve the model's robustness to diverse lighting scenarios. With the input augmentation including brightness scaling and shadow based relighting, we finally generate 200K high quality video editing pairs.

\subsubsection{3D rendered data.} 
Extracting illumination conditions from real videos inherently introduces noise. To address this limitation, we render dataset using the Cycles renderer in Blender~\cite{blender2018blender}, utilizing publicly available 3D assets from Objaverse~\cite{deitke2023objaverse} and environment maps from Poly Haven\footnote{\url{https://polyhaven.com/}}. We select 10K objects with high quality mesh from Objaverse and 1K high quality HDR environment maps. For each object placed in random selected 5 environment maps, we render five videos with random camera trajectories. After augmentation, we select one of the other four lighting conditions as the relighting input, denoted as $\mathcal{V}_{\text{rel}}$. This results in a dataset of 1M video pairs, with each pair corresponding to environment map variations for illumination modeling. This part of data containing precise HDR video maps serves as a valuable supplement to in-the-wild videos.

\subsection{Model Design}
\label{sec3.2}


Given the inherent similarities between image and video relighting tasks, we adopt a pre-trained 2D image relighting diffusion model~\cite{iclight} as our foundational model. By utilizing this model's weights for initialization, we effectively leverage its image relighting priors and accelerate the training process. However, video relighting introduces additional challenges, including dynamic illumination and motion variations. Maintaining the original relighting quality of the model, while ensuring temporal consistency across frames, are two critical issues that must be addressed.

\subsubsection{Lifting image diffusion model for video relighting.}

The overall \methodname pipeline is illustrated in \cref{fig:pipe}. Our approach to address the challenges of lifting image relighting to the video domain is as follows: First, we inflate the 2D image diffusion model into a 3D U-Net, enabling it to accept video tensor with temporal dimension as input while maintaining the core structure of the original model. To further enhance temporal consistency, we integrate temporal attention layers into the image relighting model. During training, the spatial layers are kept frozen, while only the temporal layers are fine-tuned. This strategy preserves the in-the-wild editing capability of IC-Light~\cite{iclight} and enables robust generalization to out-of-domain cases beyond the domain of our current data pair collection pipeline.

To achieve conditional illumination editing, we encode both the relighted video $\mathcal{V}_{\text{rel}}$ and the background video $\mathcal{V}_{\text{bg}}$ using a VAE encoder to obtain their latent space representations, $\mathbf{z}_{\text{rel}}$ and $\mathbf{z}_{\text{bg}}$, respectively. We then add noise for $t$ steps to obtain the noisy latent $\mathbf{z}_t$, and concatenate $\mathbf{z}_t$, $\mathbf{z}_{\text{rel}}$, and $\mathbf{z}_{\text{bg}}$ as input to the diffusion model, where $\mathbf{z}_{\text{bg}}$ serves as the background condition for relighting control. 

For environment HDR condition enjection, we encode the environment map $E$ using a 5-layer MLP, decomposing it into LDR and HDR maps $\mathbf{E}_l$ and $\mathbf{E}_h$. Textual conditions are encoded via a CLIP Text Encoder into $\mathbf{y}$, which is repeated and concatenated with the temporal HDR latents. This combined information is injected into the spatial layers via cross-attention, enabling precise control over illumination changes.

\begin{figure*}[htbp]
\centering
\includegraphics[width=1.0\textwidth]{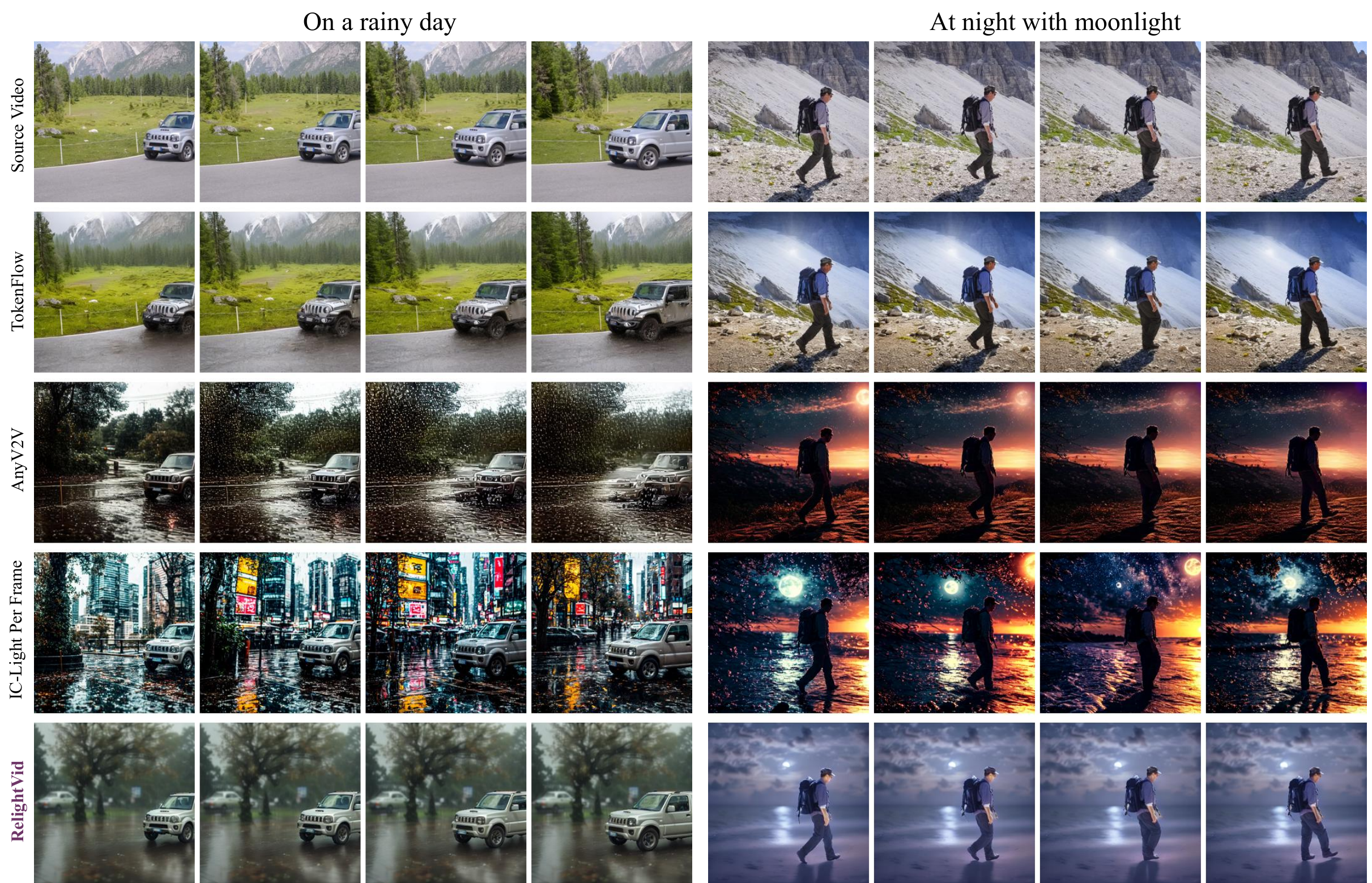}
\vspace{-6mm}
\caption{\textbf{Qualitative comparison of text-conditioned video illumination editing.} Given a source video and guidance text, we compare \methodname with other classic text-driven video editing methods, where AnyV2V initially uses ICLight to modify the first frame.}
\label{fig:text_result}
\vspace{-2mm}
\end{figure*}

\begin{table*}[htbp]
  \centering
  \scalebox{1.00}{
    \begin{tabular}{rccccccc}
    \toprule
    \multirow{2}[0]{*}{Methods} & \multicolumn{1}{c}{\multirow{2}[0]{*}{CLIP-Text↑}} & \multicolumn{1}{c}{\multirow{2}[0]{*}{CLIP-Image↑}} & \multicolumn{1}{c}{\multirow{2}[0]{*}{R-Motion Smoothness↓}} & \multicolumn{4}{c}{User Preference} \\
    \cline{5-8}
          &       &       &       & \multicolumn{1}{c}{VS↑} & \multicolumn{1}{c}{LR↑} & \multicolumn{1}{c}{TA↑} & \multicolumn{1}{c}{IP↑}\\
    \midrule
    IC-Light~\cite{iclight} Per Frame &   \textbf{0.2512}    &   0.8533   &   6.293   &   2.367   & 2.253 & 2.719 & 2.318 \\
    TokenFlow~\cite{geyer2023tokenflow} &     0.2165  &   0.9811   &   1.965    &  2.344  & 2.613 & 2.186 & 2.343 \\
    AnyV2V~\cite{ku2024anyv2v}+ICLight &  0.2372  &  0.9760  &   2.256   &   2.217    & 2.202 & 2.236 & 2.314 \\
    \methodname &    0.2493   &   \textbf{0.9841}   &   \textbf{1.683}    &  \textbf{3.072}     & \textbf{2.933} & \textbf{2.859} & \textbf{3.025} \\
    \bottomrule
    \end{tabular}%
    }
  \vspace{1mm}
  \caption{\textbf{Quantitative evaluation result of text conditioned video relighting.} \methodname achieves best result compared to other video editing methods. }
  \vspace{-8mm}
  \label{tab:text_com}%
\end{table*}%

\subsubsection{Multi-Modal Condition Joint Training.} 
Our video diffusion model is designed to enable collaborative video relighting by simultaneously leveraging both background visual condition and texture prompts for fine-grained control over illumination. The model comprises a VAE encoder $\mathcal{E}_{i}$, a denoiser 3D U-Net $\epsilon_\theta$, a CLIP-Text Encoder $\mathcal{E}_{t}$, an HDR Encoder $\mathcal{E}_{e}$, and a decoder $\mathcal{D}$. To achieve this, we introduce a joint training objective that optimizes the model for collaborative conditioning on both background and text modalities:

\begin{equation}
\min_\theta \mathbb{E}_{z \sim \mathcal{E}(x), t, \epsilon \sim \mathcal{N}(0, 1)} \| \epsilon - \epsilon_\theta (z_t, t, \hat{\mathcal{E}}) \|_2^2, 
\end{equation}
\vspace{-2.5mm} 
\begin{equation}
\hat{\mathcal{E}} = \{\mathcal{E}_{i}(z_{rel}), \mathcal{E}_{i}(z_{bg}), \mathcal{E}_{t}(y), \mathcal{E}_{e}(E)\},
\end{equation}
where $t \sim [1, 1000]$ is the diffusion time step, and $\hat{\mathcal{E}}$ represents the encoded conditional latents for the input video $V_{rel}$, background video $V_{bg}$, environment map $\hat{E}$, and the CLIP embedding of the input caption $y$. Our key innovation lies in the collaborative editing framework, which allows the model to dynamically integrate and balance information from both image background and text prompts during the editing process. This enables precise and coherent video editing where background and text conditions work synergistically to guide the output, ensuring that the edited video aligns with both visual and textual context.

\subsubsection{Illumination-Invariant Ensemble.}  
\begin{figure*}[htbp]
\centering
\includegraphics[width=0.97\textwidth]{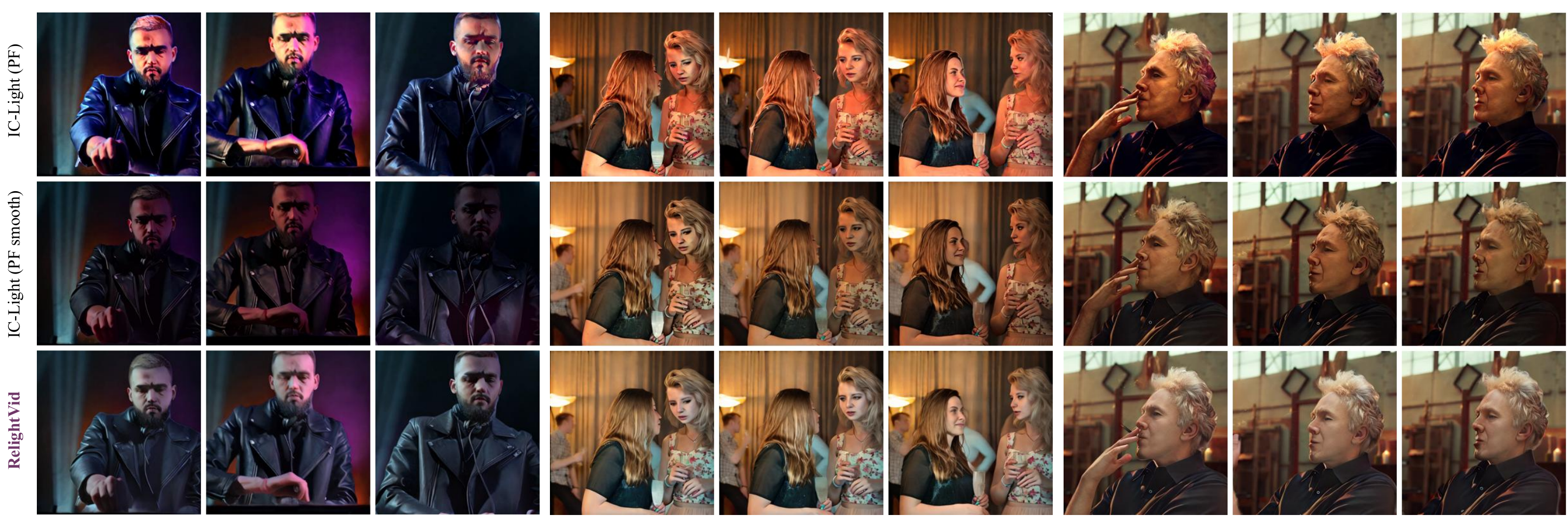}
\vspace{-5mm}
\caption{\textbf{Qualitative comparison of background-conditioned video illumination editing.} Given any foreground apperance and a background video reference, we relight videos and compare our method with the per-frame IC-Light (smoothed) method.}
\vspace{-3mm}
\label{fig:bg_result}
\end{figure*}

\begin{table*}[htbp]
  \centering
  \scalebox{0.93}{
    \begin{tabular}{rcccccc}
    \toprule
    \multirow{2}{*}{Methods} & \multirow{2}{*}{PSNR↑} & \multirow{2}{*}{SSIM↑} & \multirow{2}{*}{LPIPS↓} & \multirow{2}{*}{R-Motion Smoothness↓} & \multicolumn{2}{c}{User Preference} \\
    \cline{6-7}
    & & & & & Video Smoothness↑ & Lighting Rationality↑ \\
    \midrule
    IC-Light~\cite{iclight} Per Frame &   18.26    &    0.7608   & 0.1533 & 2.508        &   1.849    &    1.892   \\
    Per Frame + Smoothing &    17.71   &    0.7626   & 0.1732 & 2.383     &  1.836  &   1.929  \\
    \methodname &   \textbf{18.79}    &   \textbf{0.7832}    &   \textbf{0.1412}  &   \textbf{1.458}   &  \textbf{2.315}  & \textbf{2.180} \\
    \bottomrule
    \end{tabular}%
    }
  \caption{\textbf{Quantitative evaluation results of background video conditioned relighting.} \methodname maintains the image relighting ability of IC-Light and significantly improve temporal consistency.}
  \vspace{-6mm}
  \label{tab:video_bg_com}%
\end{table*}

In a single relighting process given background video condition, background video $\mathcal{V}_{\text{bg}}$ is fixed, Consequently, the relighted foreground $\mathcal{V}_{\text{rel}}$ should ideally be fixed. This ideal output should remain the same regardless of any brightness augmentation applied to the original input. Motivated by this observation, we propose an Illumination-Invariant Ensemble (IIE) strategy to enhance the robustness of video relighting.
The core idea behind IIE is to apply brightness augmentations to the original input video and then average the predicted noise to obtain a more reliable result. The rationale is that different augmented inputs should guide the noisy latent toward the same output video, thereby mitigating the impact of illumination variations.

Specifically, We first apply a series of brightness augmentations directly to the input video $\mathcal{V}_{\text{in}}$, generating $N$ augmented versions. The augmented video frames are defined as:  

\begin{equation}
\mathcal{V}_{\text{in}}^{(i)} = s_i \cdot \mathcal{V}_{\text{in}}, \quad i = 1, 2, \dots, N,
\end{equation}  

Where $\{s_i\}_{i=1}^N$ are brightness scaling factors sampled from a predefined range, e.g., $s_i \in [0.5, 1.5]$. These augmented videos are then fed into the model to predict the noise $\mathbf{\epsilon}_t^{(i)}$ at each diffusion step $t$. To obtain a more robust denoising result, we compute the averaged noise prediction across all augmented versions: $\mathbf{\bar{\epsilon}}_t = \frac{1}{N} \sum_{i=1}^N \mathbf{\epsilon}_t^{(i)}$. Finally, the averaged noise $\mathbf{\bar{\epsilon}}_t$ is used in the diffusion process to produce the final relighted video output. This strategy effectively improves the model's robustness under varying illumination conditions, preventing undesirable variations in albedo. The effectiveness of IIE is validated in \cref{sec:iie}.

\label{sec3.3}

\section{Experiments}

\subsection{Training Details}
We adopt the SD-1.5~\cite{rombach2022high} version of IC-Light~\cite{iclight} as the image backbone and inject temporal attention layers initialized from AnimateDiff-V2~\cite{guo2023animatediff}. For the HDR environment map encoder, we initialize its parameters with zeros to minimize its influence at the beginning of training. During training, the cross-attention layers between image features and HDR features, as well as the temporal layers, are made trainable, while the other parts of the UNet are kept fixed. The learning rate is set to $1\times10^{-5}$ with AdamW~\cite{loshchilov2017decoupled} optimizer adopted. Training is conducted on 8 NVIDIA A100-80G GPUs for 5,000 iterations.


\begin{figure}[htbp]
\centering
\includegraphics[width=0.45\textwidth]{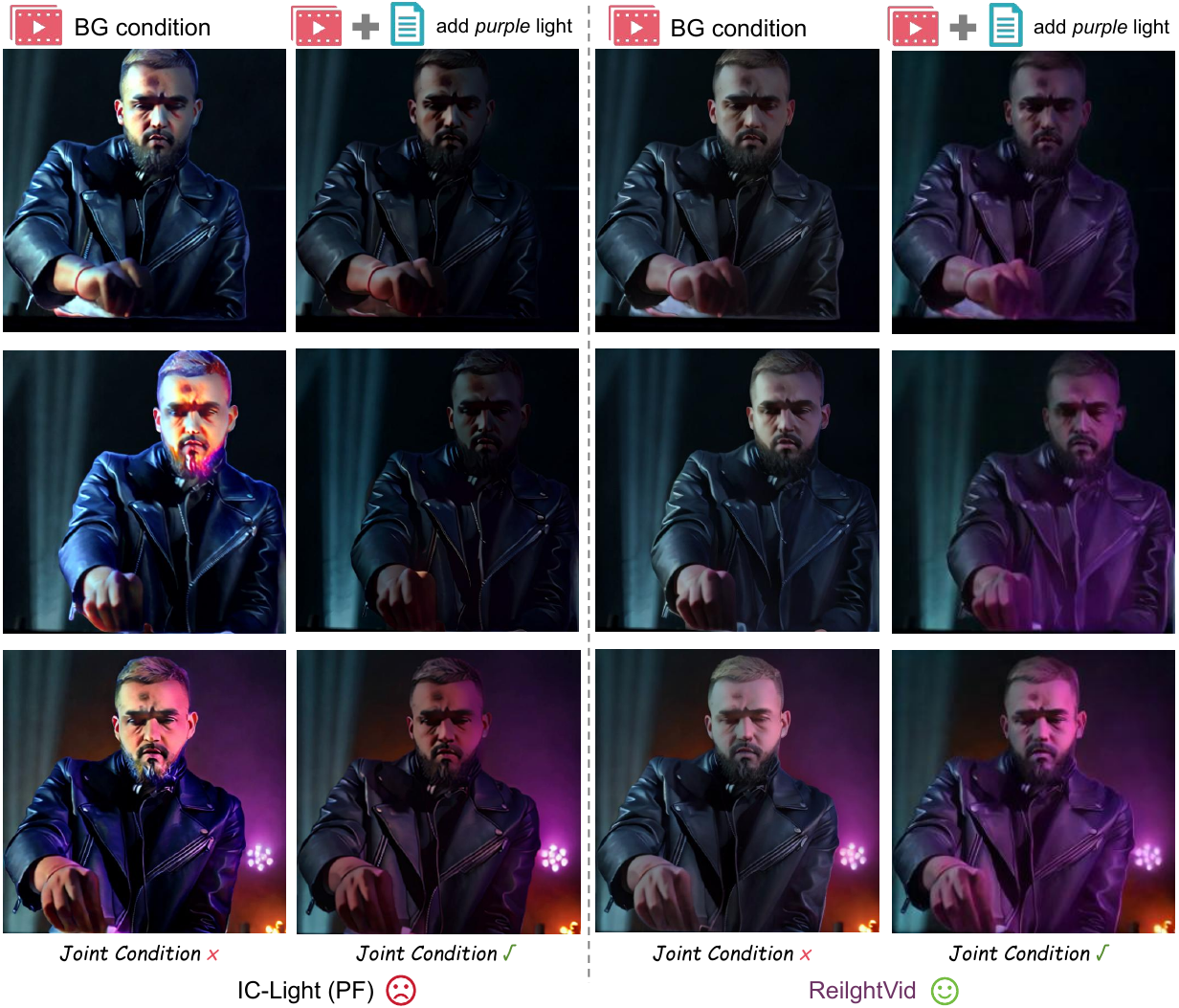}
\vspace{-3mm}
\caption{\textbf{Synthetic background-condition illumination editing results.} We use the Hunyuan model for long synthetic background videos with strong dynamic lighting, demonstrating the effectiveness and robustness of our method in scenarios with dynamic lighting and long videos editing.}
\label{fig:bg_sync_result}
\vspace{-2mm}
\end{figure}

\subsection{Evaluations}

\begin{figure}[htbp]
\centering
\includegraphics[width=0.48\textwidth]{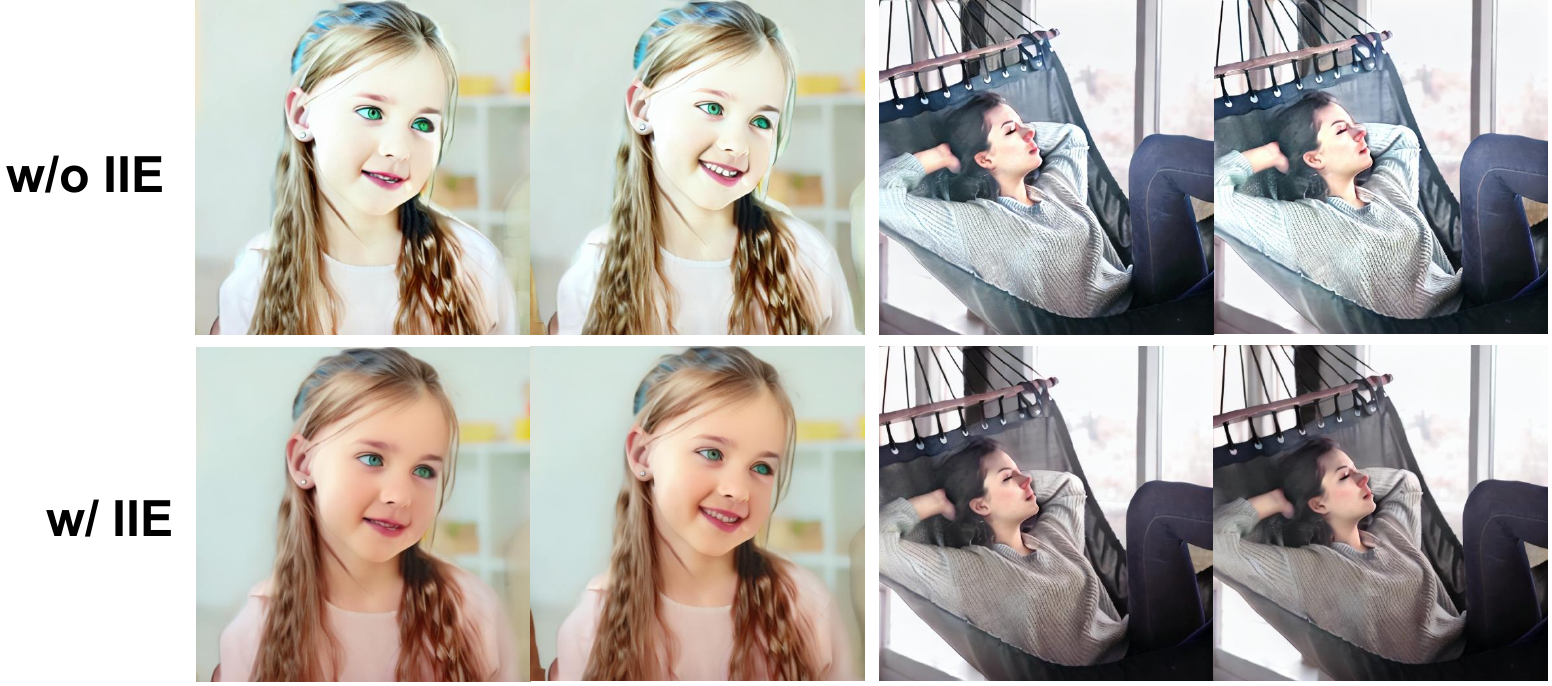}
\vspace{-7mm}
\caption{\textbf{Qualitative results of Illumination-Invariant Ensemble.} IIE helps produce more robust result that better preserve albedo in light editing.}
\label{fig:iie_result}
\vspace{-3mm}
\end{figure}

\begin{figure}[htbp]
\centering
\includegraphics[width=0.47\textwidth]{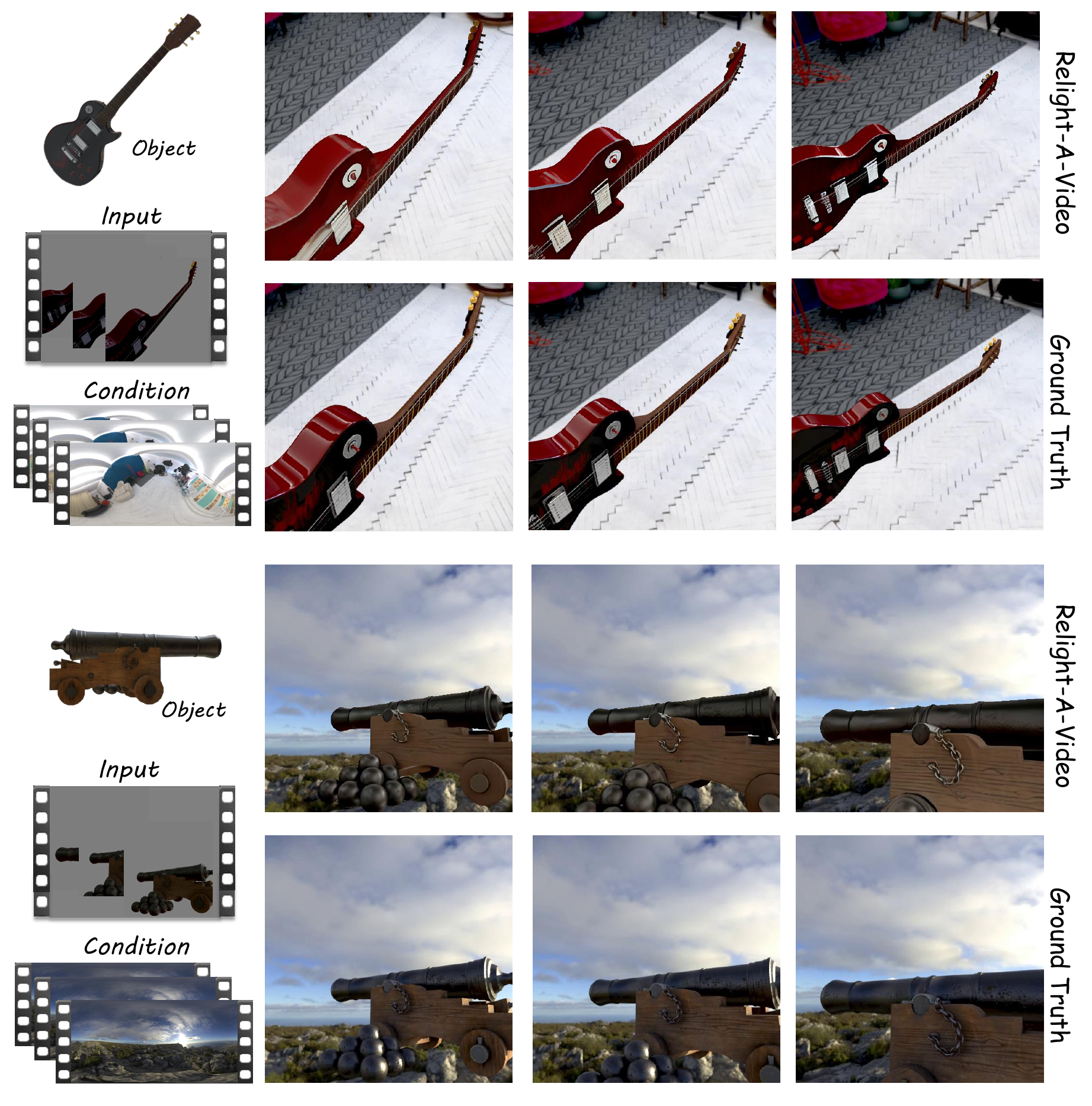}
\vspace{-5mm}
\caption{\textbf{HDR-conditioned illumination editing results.}}
\label{fig:hdr_result}
\vspace{-4mm}
\end{figure}

\subsubsection{Video Background Conditioned Relighting.} 
\label{sec:vbc_relighting}
To assess both image quality and temporal consistency of \methodname, we sample 50 in-the-wild videos from the Mixkit dataset, encompassing both static and dynamic lighting conditions. The foreground portrait or object is augmented with diverse relighting transformations, serving as model input for evaluation. We setup a baseline of per-frame relighting via IC-Light~\cite{iclight} with the video smoothing method. For image quality, we adopt PSNR, SSIM~\cite{wang2004image}, LPIPS~\cite{zhang2018unreasonable} as evaluation metrics. For temporal consistency, we adopt Motion Smoothness metric proposed in~\cite{huang2024vbench} through utilize the motion priors in video frame interpolation model~\cite{li2023amt}. We conducted user study comparing our methods and baseline methods across two dimensions: Video Smoothness ( The consistency between different frames), and Lighting Rationality (The rationality of lighting on foreground object/human). We invite 41 users including graduate students that expertise in video editing and average users to rank the results and use Average User Ranking (AUR) as a preference metric.

As shown in \cref{tab:video_bg_com}, \methodname consistently outperforms all other methods across all evaluated metrics. It not only significantly improves video smoothness but also preserves and subtly enhances the relighting capabilities of IC-Light~\cite{iclight}, while intrinsically maintaining the foreground albedo. Qualitative comparisons are provided in \cref{fig:bg_result}.

\subsubsection{Text-Conditioned Relighting.}
Text-conditioned video relighting represents a versatile setting with wide-ranging applications for real-world users. We compare our method with per-frame IC-Light~\cite{iclight} and two state-of-the-art video editing methods: TokenFlow~\cite{geyer2023tokenflow} and AnyV2V~\cite{ku2024anyv2v}. TokenFlow directly leverages text prompts for video editing, while AnyV2V conditions on the original video and the first edited frame. For AnyV2V, we use IC-Light~\cite{iclight} to relight the first frame. The evaluation is conducted on the DAVIS dataset~\cite{perazzi2016benchmark}, where we randomly sample 10 text relighting prompts with corresponding relighting guidance for each video. We measure text-to-video alignment using the CLIP-Text score and evaluate semantic consistency across frames using the CLIP-Image score.

We conduct user study comparing our method with baselines across four dimensions: \textbf{V}ideo \textbf{S}moothness, \textbf{L}ighting \textbf{R}ationality (as defined in \cref{sec:vbc_relighting}), and two new metrics: \textbf{T}ext \textbf{A}lignment (alignment between video content and text prompt) and \textbf{I}D-\textbf{P}reser-vation (consistency of the foreground object’s identity and albedo before and after relighting). A total of 37 participants similar to \cref{sec:vbc_relighting} are invited to rank the results.

As reported in \cref{tab:text_com}, \methodname generally outperforms baseline approaches and state-of-the-art video editing methods. Representative results are visualized in \cref{fig:text_result}. \methodname demonstrates the ability to perform high-quality relighting that faithfully adheres to the given text prompts while preserving the identity and albedo of the foreground object or human.

\begin{table}[t]
  \centering
    \begin{tabular}{r|ccc}
    \toprule
          & PSNR  & SSIM & R-Motion Smoothness\\
    \midrule
    Single Input & 18.79 & 0.7832 & 1.458 \\
    3-Aug-Input & 18.92 & 0.7907 & 1.449 \\
    5-Aug-Input & 19.07 & 0.7904 & 1.452 \\
    \bottomrule
    \end{tabular}%
    \vspace{1mm}
    \caption{\textbf{Quantitative results of Illumination Invariant Ensemble.}}
    \vspace{-8mm}
  \label{tab:iie}%
\end{table}%

\subsubsection{HDR-Conditioned Relighting.}
In addition to background and textual cues as relighting conditions, we also incorporate HDR video as a more precise condition for relighting. As demonstrated in \cref{fig:hdr_result}, our model, by injecting an HDR map through a specially designed HDR encoder and leveraging cross-attention interactions with frame features, is able to recover most of the relevant information and apply accurate relighting for the foreground object. This enhanced control is particularly effective in scenarios where the predominant lighting originates from the camera's direction, where background and text-based cues struggle to address.

\subsubsection{Illumination-Invariant Ensemble (IIE)}
\label{sec:iie}
We evaluate the performance of the Illumination-Invariant Ensemble (IIE) using the same test set as in the video background-conditioned relighting task, with two augmented foreground videos and the original video as inputs. As demonstrated in \cref{fig:iie_result} and summarized in \cref{tab:iie}, the incorporation of IIE significantly enhances the robustness of relighting and improves the preservation of foreground albedo, such as the shirt of the girl and the hammock. However, an increase in the number of augmented inputs may result in an average blurring effect, which can negatively impact the overall performance.





\section{Conclusion}

In this work, we propose \methodname, a video diffusion model that supports relighting any foreground object in a video conditioned on a new background video, text, and an HDR map without requiring complex process of intrinsic decomposition. We demonstrate its promising performance across various scenarios and explore key factors behind its success: a carefully designed data generation pipeline and the efficient reuse of prior knowledge from the image backbone. We believe that this model can serve as a versatile tool to fulfill real user requirements and hope this work will inspire future research on video diffusion models for editing and generation.




\newpage
\bibliographystyle{ACM-Reference-Format}
\bibliography{main}

\appendix

\begin{figure*}[htbp]
\centering
\includegraphics[width=0.97\textwidth]{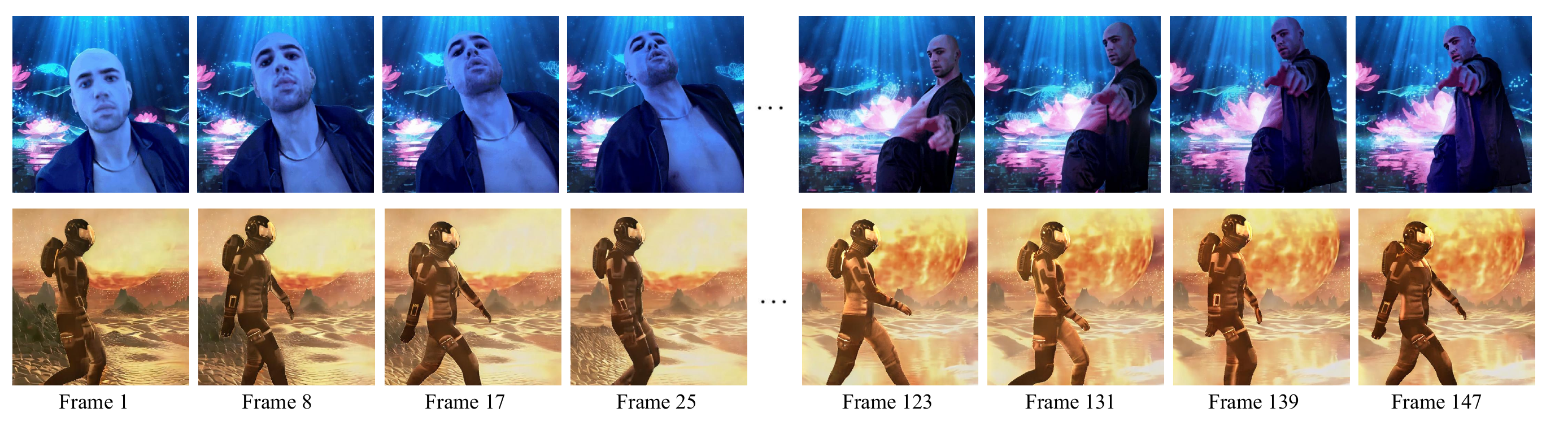}
\vspace{0mm}
\caption{\textbf{Qualitative results of background-conditioned long video illumination editing.} Given any length foreground apperance video and a background video reference, we relight videos using long video propagation method.}
\vspace{4mm}
\label{fig:bg_long}
\end{figure*}


\begin{figure*}[htbp]
\centering
\includegraphics[width=0.99\textwidth]{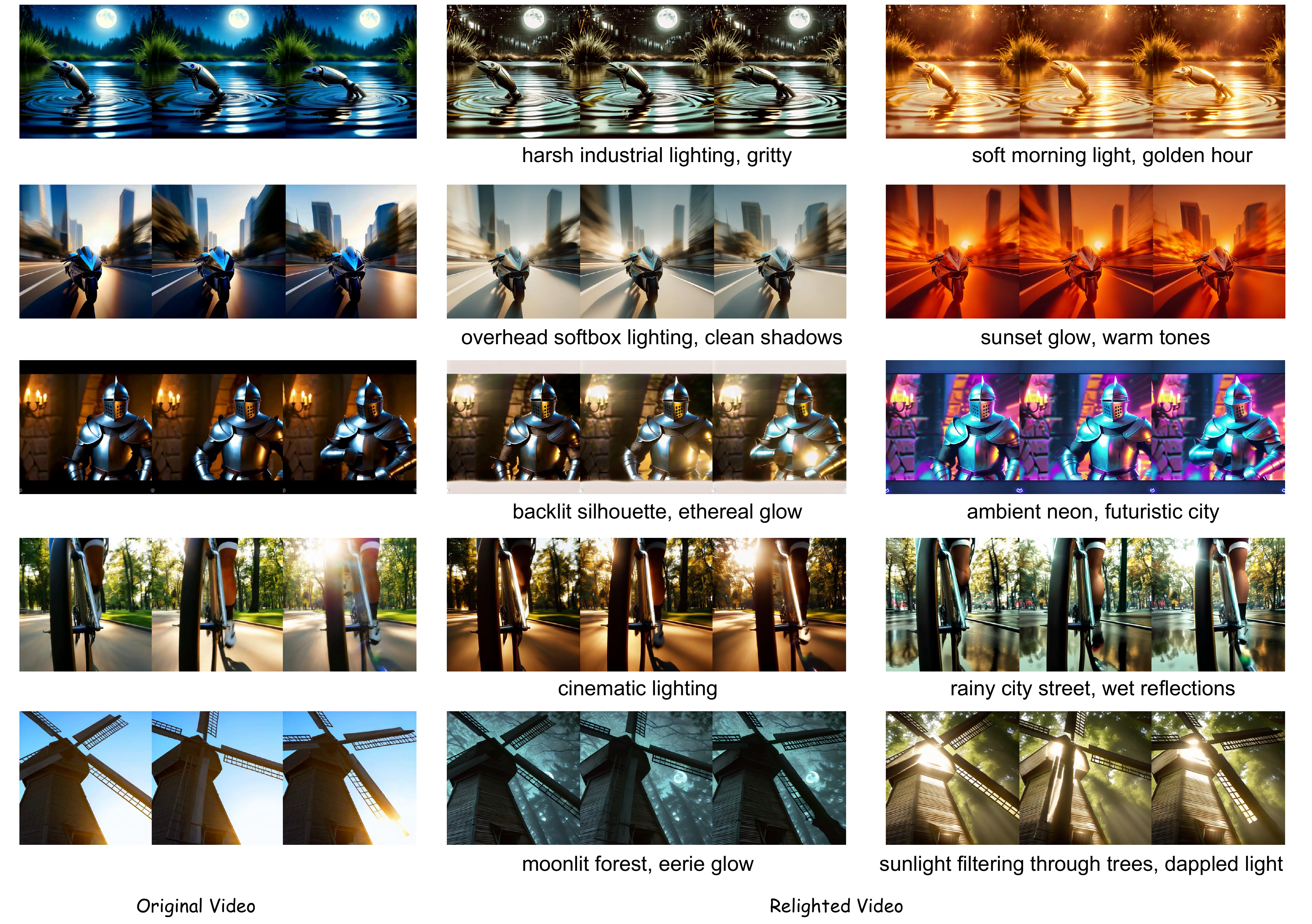}
\vspace{0mm}
\caption{\textbf{More qualitative results of text-conditioned video illumination editing beyond test set.}}
\vspace{0mm}
\label{fig:text_more}
\end{figure*}

\begin{figure*}[htbp]
\centering
\includegraphics[width=0.95\textwidth]{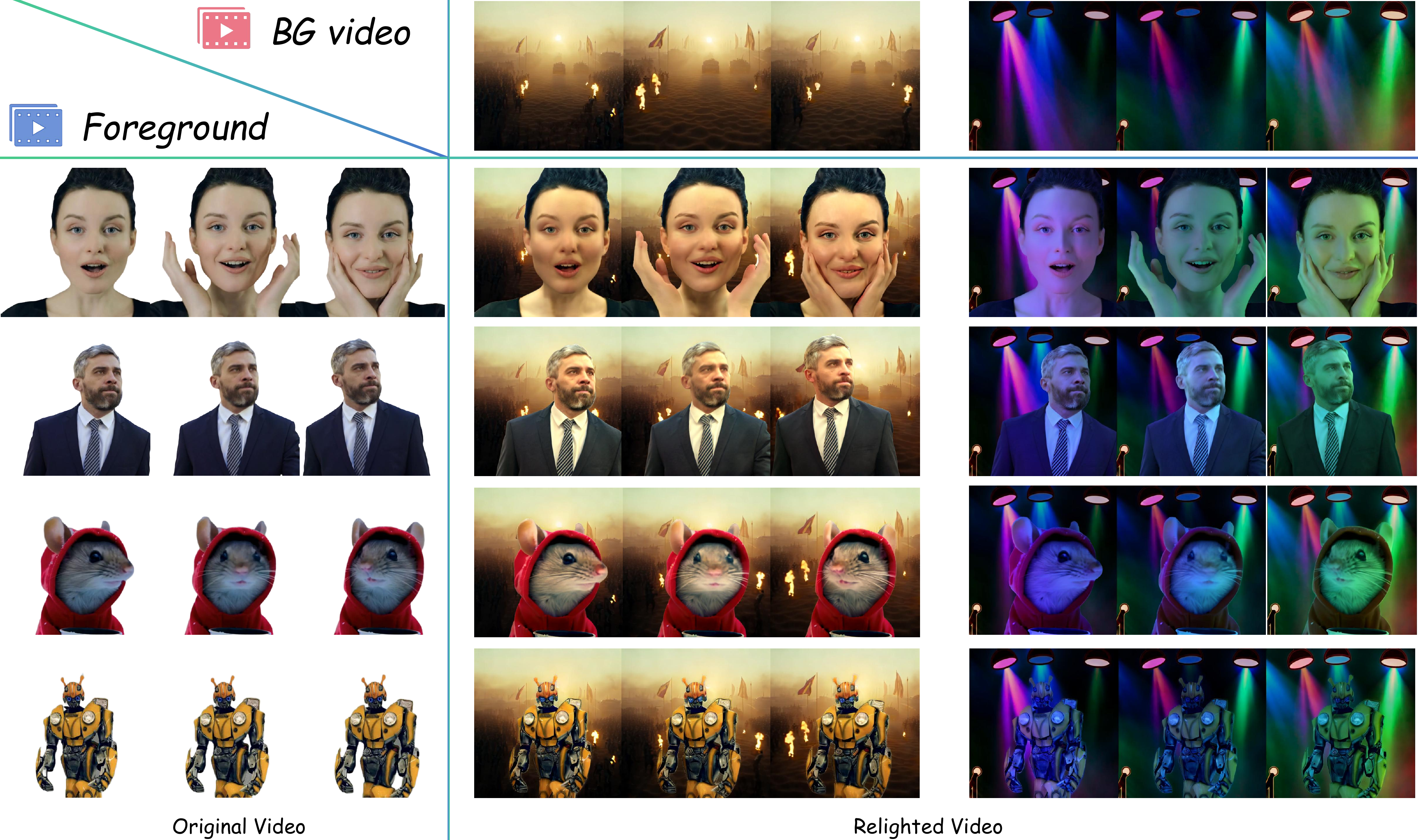}
\vspace{0mm}
\caption{\textbf{More qualitative results of background-conditioned video illumination editing beyond test set.}}
\vspace{0mm}
\label{fig:bg_more}
\end{figure*}

\begin{figure*}[htbp]
\centering
\includegraphics[width=0.94\textwidth]{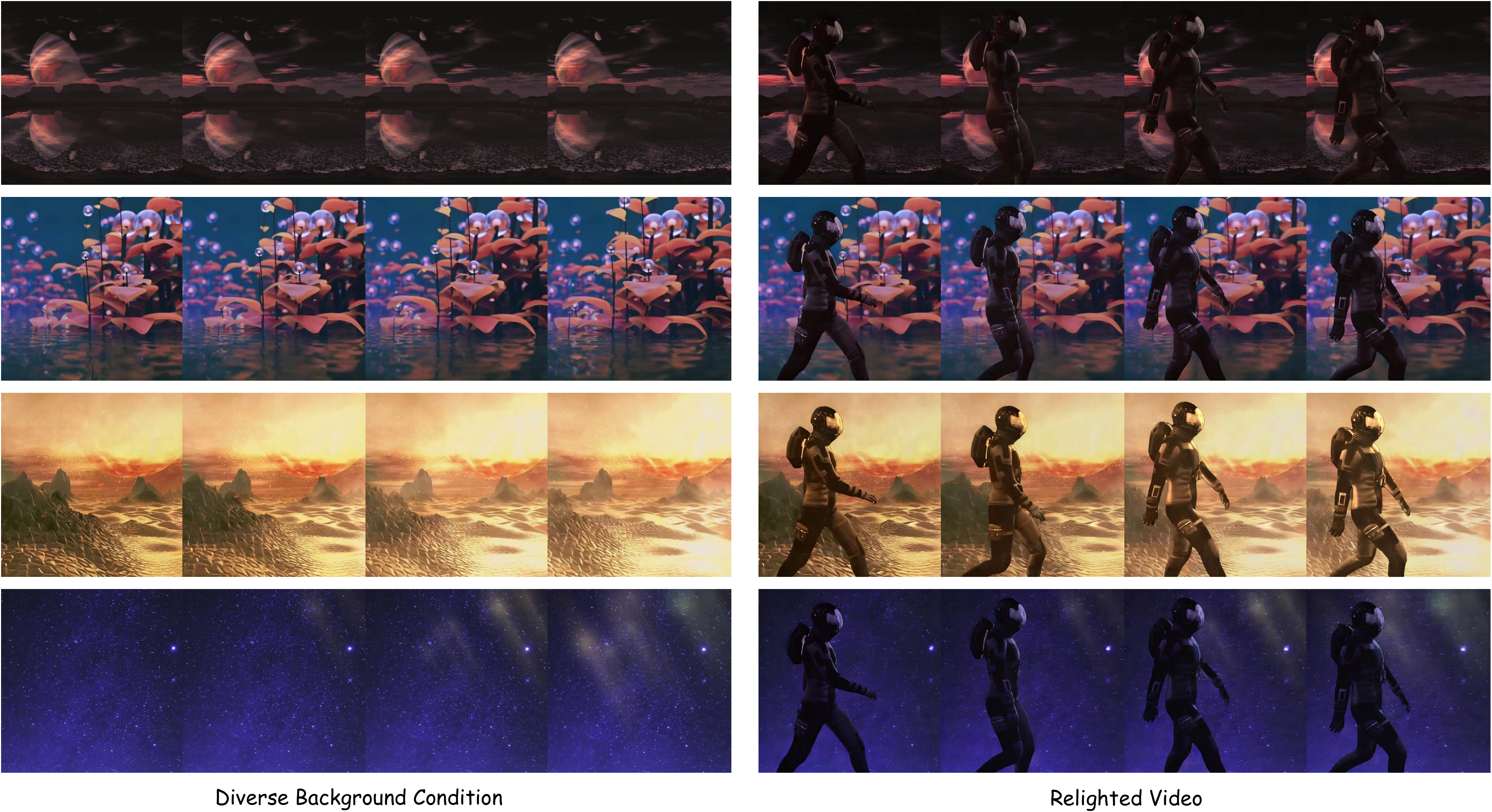}
\vspace{0mm}
\caption{\textbf{More qualitative results of background-conditioned video illumination editing beyond test set.}}
\vspace{0mm}
\label{fig:bg_more2}
\end{figure*}

\end{document}